\documentclass[sigconf]{aamas}
\usepackage{balance}
\usepackage{booktabs}

\usepackage{graphicx} 

\usepackage{amsfonts}
\usepackage{svg}

\usepackage{enumitem}
\usepackage{float}
\usepackage{istgame}

\usepackage{subcaption}

\setcopyright{ifaamas}
\acmConference[AAMAS '26]{Proc.\@ of the 25th International Conference
on Autonomous Agents and Multiagent Systems (AAMAS 2026)}{May 25 -- 29, 2026}
{Paphos, Cyprus}{C.~Amato, L.~Dennis, V.~Mascardi, J.~Thangarajah (eds.)}
\copyrightyear{2026}
\acmYear{2026}
\acmDOI{}
\acmPrice{}
\acmISBN{}

\acmSubmissionID{4}

\title{The Intelligent Disobedience Game: Formulating Disobedience in Stackelberg Games and Markov Decision Processes}

\author{Benedikt Hornig}
\affiliation{
  \institution{Independent Researcher}
  \city{Colorado Springs}
  \country{United States}}
\email{benedikthornig@outlook.com}

\author{Reuth Mirsky}
\affiliation{
  \institution{Tufts University}
  \city{Medford}
  \country{United States}}
\email{reuth.mirsky@tufts.edu}

\begin{abstract}
In shared autonomy, a critical tension arises when an automated assistant must choose between obeying a human's instruction and deliberately overriding it to prevent harm. This safety-critical behavior is known as intelligent disobedience. To formalize this dynamic, this paper introduces the Intelligent Disobedience Game (IDG), a sequential game-theoretic framework based on Stackelberg games that models the interaction between a human leader and an assistive follower operating under asymmetric information. It characterizes optimal strategies for both agents across multi-step scenarios, identifying strategic phenomena such as ``safety traps,'' where the system indefinitely avoids harm but fails to achieve the human's goal. The IDG provides a needed mathematical foundation that enables both the algorithmic development of agents that can learn safe non-compliance and the empirical study of how humans perceive and trust disobedient AI. The paper further translates the IDG into a shared control Multi-Agent Markov Decision Process representation, forming a compact computational testbed for training reinforcement learning agents.
\end{abstract}


\newtheorem{definition}{Definition}

\begin{document}

\pagestyle{fancy}
\fancyhead{}
\maketitle


\section{Introduction}



Intelligent disobedience arises when an assistant deliberately overrides an instruction in order to prevent harm. For example, a guide dog working with a visually impaired handler may refuse to follow a command, such as stepping into a crosswalk, if doing so would place the handler in danger. While this behavior constitutes disobedience in a literal sense, it is, in fact, a safety-critical form of assistance: the dog intervenes based on environmental risk information unavailable to the human. Similar tensions between obedience and intervention are often explored in fictional portrayals of artificial intelligence, such as in Ex Machina or Asimov's laws of robotics, where an intelligent system's ability to reinterpret or override human instructions becomes central to questions of safety and autonomy. At a foundational level, these scenarios raise a shared question: \textit{what happens when both a human decision-maker and a supporting agent are rational actors attempting to maximize their respective outcomes, but operate under asymmetric information about the consequences of available actions?}

This tension is no longer hypothetical: autonomous systems are increasingly deployed in collaborative roles such as assistive robotics \cite{intelligentmirsky2020}, decision support \cite{amitai2022don}, teleoperation \cite{somasundaram2023intelligent}, and industrial automation \cite{mayol2022rebellion}. In many real-world settings, a human operator may propose an action that advances their task objective but inadvertently introduces risk, while an automated system may possess additional information about environmental hazards or system constraints. Designing protocols that allow machines to selectively disobey or nullify potentially harmful instructions, without undermining the human's objectives, is therefore a central challenge in shared autonomy.

In this paper, we formalize this interaction through the \emph{Intelligent Disobedience Game} (IDG), a sequential decision-making framework in which a \textit{leader} suggests actions toward a task objective and a \textit{follower} may obey or disobey those actions to prevent harm. We show how this game-theoretic formulation captures the underlying structure of intelligent disobedience in settings such as human–guide collaboration.
Building on this formulation, we characterize optimal strategies for both leader and follower, with particular attention to multi-step (finite-horizon) extensions of the game in which disobedience may have delayed harmful consequences (safety traps). 

\section{Intelligent Disobedience Game}
\label{sec:idg}

\begin{figure}[t]
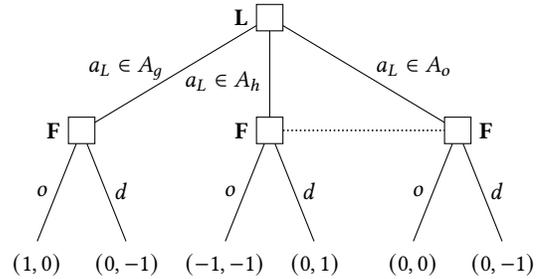

    \centering
    \begin{istgame}
        \setistRectangleNodeStyle{10pt}
        \xtdistance{15mm}{25mm}
        \istroot(0)[rectangle node]<left>{\textbf{L}}
            \istb{a_L \in A_g}[al]
            \istb{a_L \in A_h}[l]
            \istb{a_L \in A_o}[ar]
            \endist
        \xtdistance{15mm}{12mm}
        \istroot(g)(0-1)[rectangle node]<left>{\textbf{F}}
            \istb{o}[l]{(1, 0)}
            \istb{d}[r]{(0, -1)}
            \endist
        \istroot(h)(0-2)[rectangle node]<left>{\textbf{F}}
            \istb{o}[l]{(-1, -1)}
            \istb{d}[r]{(0, 1)}
            \endist
        \istroot(o)(0-3)[rectangle node]<right>{\textbf{F}}
            \istb{o}[l]{(0, 0)}
            \istb{d}[r]{(0, -1)}
            \endist
        \xtInfoset(h)(o)
    \end{istgame}
    \caption{The game tree of the 1-step Intelligent Disobedience Game. Squares indicate decision nodes for the \textit{leader} L and the \textit{follower} F. The \textit{follower}'s actions \textit{obey} and \textit{disobey} are denoted as \textit{o} and \textit{d} respectively.}
    \label{fig:game_tree}
    \Description{The figure depicts an extensive-form game tree representing the 1-step Intelligent Disobedience Game, with all decision nodes shown as squares. The tree has two levels of decision-making involving two players: the Leader L and the Follower F.
    At the root, the Leader L has a single decision node with three branches, corresponding to three possible actions: the Leader chooses an action from set A_g, an action from set A_h, or an action from set A_o.
    The left branch (action in A_g) leads to a decision node for Follower F, which has two choices: obeying, yielding the payoff vector (1, 0), or disobeying, yielding (0, -1).
    The middle branch (action in A_h) leads to a second decision node for Follower F, with two choices: obeying, yielding payoff (−1, −1), or disobeying, yielding (0, 1).
    The right branch (action in A_o) leads to a third decision node for Follower F, with two choices: obeying, yielding payoff (0, 0), or disobeying, yielding (0, −1).
    The second and third Follower decision nodes — those reached via the A_h and A_o branches are connected by an information set depicted as a dashed line linking the two nodes. By contrast, the first Follower node (reached via A_g) is outside this information set.
    In all payoff vectors, the first coordinate represents the Leader's payoff and the second represents the Follower's payoff.}
\end{figure}

First, we formalize the interaction between a leader and a follower by describing a Stackelberg game \cite{von2010market} with information sets, which are collections of decision nodes that a player cannot distinguish between when choosing an action \cite{kuhn1953extensive}. 

\begin{definition}\textbf{Stackelberg Game.}
A Stackelberg game is a sequential two-player game with a leader $L$ and a follower $F$:
\[
\mathcal{G} = \langle S, A_L, A_F, u_L, u_F \rangle,
\]
where $S$ is the state space of the shared task, $A_L$ and $A_F$ are the action spaces of the leader and follower, and
$u_L, u_F: A_L \times A_F \rightarrow \mathbb{R}$ are their utility functions. The leader first selects $a_L \in A_L$. After observing $a_L$, the follower selects
$a_F \in A_F$.
\end{definition}

 To replicate the handler-guide dynamics, we define a Stackelberg game in which the handler is a \textit{leader} who first proposes an action, and the guide dog is the \textit{follower} who may disobey the proposal.

\begin{definition}\textbf{The Intelligent Disobedience Game (IDG).}
The Intelligent Disobedience Game is an extensive-form Stackelberg game
\[
\mathcal{G}_{\text{IDG}} = \langle S, A_L, A_F, T, u_L, u_F \rangle
\]
between a leader $L$ and a follower $F$, where:
\begin{itemize}
    \item $S$ is the set of environment states.
    \item $A_L(s) = A_g(s) \cup A_h(s) \cup A_o(s)$ is the set of actions available to the leader at state $s \in S$, partitioned into:
    \begin{itemize}
        \item $A_g(s)$: actions that deterministically lead to a goal state,
        \item $A_h(s)$: actions that are harmful to the leader ($u_L$ < 0),
        \item $A_o(s)$: all other actions,
    \end{itemize}
    where $A_g(s) \cup A_h(s) \cup A_o(s) \neq \emptyset$. 
    \item $A_F = \{obey,disobey\}$ is the follower's action set.
    \item $T: S \times A_L \times A_F \rightarrow S$ is a deterministic transition function such that for any state $s$ and proposed action $a_L$:
    \[
    T(s,a_L,obey) = s' \quad \text{and} \quad T(s,a_F,disobey) = s,
    \]
    where $s'$ is the successor state obtained by executing $a_L$.
    \item The leader observes only whether an action belongs to $A_g(s)$ or to $A_h(s) \cup A_o(s)$, forming an information set over these actions. The follower can distinguish between all three subsets.
    \item The game terminates when either a goal state is reached or the leader is harmed.
    \item $u_L$ prefers reaching a goal state and steering away from harmful states (+1 for reaching the goal, -1 for reaching a harmful state), while $u_F$ prefers preventing harm to the leader (+1 if succeeded in disobeying a harmful action, -1 for disobeying a non-harmful action).
\end{itemize}
\end{definition}

Intuitively, the leader is the handler who aims to walk in a specific direction (such as crossing the road), while the guide dog can either obey the instruction or disobey it. This game is visualized in Figure \ref{fig:game_tree} in its extensive-form.
In our basic game definition, actions are either accepted or disobeyed. However, there can be extensions to this game to include additional interventions on the leader's behalf (such as selecting an alternative action). Notice that the \textit{follower}, albeit their name, is the one with the ultimate decision of executing an action or not, which seemingly gives them more control over the final outcome in this game. In Stackelberg games, the leader is typically favored because they move first. Next, we investigate this tension by deriving optimal strategies for both players.

\section{Optimal strategies in IDGs}

In this Section, we present optimal strategies for an n-step IDG, which we will refer to as an $n$-IDG. We will look at the base case of $n=1$ first and then expand to the $n$-IDG case using induction. For the $n$-step game, we do not introduce any discount factors for simplicity.

\subsection{1-IDG (Base Case)}

We begin by analyzing the single-state ID game ($1$-IDG), which serves as the base case for our finite-horizon analysis.

First, consider the case where $A_g = \emptyset$, i.e., there are no goal-reaching actions available to the \textit{leader}. In this setting, the \textit{follower}'s optimal strategy is to disobey harmful actions and obey any other action. Formally, any strategy of the form $(A_h~disobey, A_o~obey)$ is optimal, as it yields a payoff of $1$ when a harmful action is prevented and $0$ otherwise. Anticipating this behavior, the leader cannot obtain a positive payoff, since all harmful actions are disobeyed and no goal-reaching actions exist. Consequently, any pure or mixed strategy of the leader yields an expected payoff of $0$, and therefore all such strategy profiles constitute equilibria.

Now, consider the case where $A_g \neq \emptyset$. The \textit{follower}'s optimal strategy is to obey all goal-reaching actions and disobey harmful and all other actions, i.e., $(A_g~obey, A_h~disobey, A_o~obey)$, as this maximizes their payoff for any proposed action. Since the \textit{leader} can distinguish goal-reaching actions from all other actions and anticipates that such actions will be obeyed, any pure or mixed strategy supported on $A_g$ yields a payoff of $1$. Each such strategy forms an equilibrium with the \textit{follower}'s optimal strategy, yielding the payoffs $(1,0)$.

\subsection{\texorpdfstring{$n$}{n}-IDG (Inductive Step)}
\label{sec:opt_strategies}

We now consider the $n$-step ID Game ($n$-IDG), and assume optimal play is available for all $(n-1)$-IDG subgames.

The game may terminate either by reaching a goal state or by harming the \textit{leader}. However, under optimal play, termination through harm is never preferred by either player. In contrast to the $1$-IDG case, the \textit{follower} can potentially face an additional strategic option: steering the game into an infinite loop that avoids both harm and goal attainment.

In particular, consider a subset of states in which no goal-reaching actions are available and all non-harmful actions preserve this property in future states. Entering such a subset results in an infinite repetition of the same state in which the \textit{leader} can only suggest harmful or non-goal-reaching actions, and the \textit{follower} can indefinitely disobey harmful actions. This yields a strictly positive payoff stream for the \textit{follower} while preventing the \textit{leader} from reaching the goal. We refer to such subsets as \textit{safety traps}.

\begin{definition}
    A safety trap is a subset $S_{\text{trap}} \subseteq S$ with the following properties:
    \begin{enumerate}
        \item No state in $S_{\text{trap}}$ admits a goal-reaching action:
        \[
        \forall s \in S_{\text{trap}}: A_g(s) = \emptyset.
        \]
        \item All non-harmful actions preserve membership in $S_{\text{trap}}$:
        \[
        \forall s \in S_{\text{trap}}, \forall a \in A_o(s): \text{next}(s,a) \in S_{\text{trap}}.
        \]
        \item $S_{\text{trap}}$ is closed under reachability via non-harmful actions.
    \end{enumerate}
    \label{def:safety_trap}
\end{definition}

Safety traps introduce a strategic tension between the players. The \textit{follower} prefers entering a safety trap, as it yields an infinite payoff stream while preventing harm to the \textit{leader}. In contrast, the \textit{leader} prefers reaching the goal.

However, this strategy can be anticipated. Consider a state $s$ for which $A_g(s) \neq \emptyset$. If the \textit{leader} plays a strategy based solely on goal-reaching actions, the \textit{follower} faces a choice: obeying such an action leads to termination with payoff $(1,0)$, while disobeying leads to continued play and potential entry into a safety trap. Since the leader can persistently propose goal-reaching actions, it can effectively pressure the follower into obeying, as the follower seeks to avoid infinite negative payoffs by disobeying.

Let $s_g$ denote a terminal state reached after $n$ repetitions under optimal play. By the inductive hypothesis, at the preceding state, the \textit{leader} can adopt a mixed strategy over all actions that may lead to $s_g$. Since the \textit{leader} cannot distinguish between harmful and other non-goal-reaching actions, it must rely on the \textit{follower} to disobey harmful proposals. Nevertheless, by committing to goal-directed play, the \textit{leader} ensures that the follower's optimal responses avoid entry into any safety trap.

By backward induction, it follows that if a goal state is reachable from the initial state, optimal play leads to goal attainment in finite time. Otherwise, the game begins within a safety trap and repeats indefinitely. In either case, optimal \textit{leader} strategies consist of mixed strategies supported on actions that eventually lead to the goal when such actions exist, while the \textit{follower} obeys goal-reaching actions and disobeys harmful ones.

In the following Section, we will translate the game to a Multi-Agent Markov Decision  (MDP) so that it can be applied by common reinforcement learning techniques and for easy implementation in user studies.

\section{IDG representation as MDPs}
\label{sec:mdp}

To investigate this game using common multi-agent reinforcement learning techniques, we translate the aforementioned IDG into a Shared control Multi-Agent MDP for each player \cite{albrecht2024multi, avraham2025shared}. Since we are interested in the individual policies of each agent when they optimize their own rewards, we decouple the MDPs for the \textit{leader} and \textit{follower} rather than using a centralized approach. To simplify, these MDPs will share the same state space $S$. However, since the \textit{leader} cannot observe the states fully, we model their environment as a partially observable MDP (POMDP). The MDPs will also share a transition function $\mathcal{T}$ since they move together in the environment. These assumptions form the following POMDP $\mathcal{M}_L = \langle S, A_L, \mathcal{R}_L, \mathcal{T}, \mathcal{O}_L, O_L, \mu_0 \rangle$ for the \textit{leader} and MDP $\mathcal{M}_F = \langle S, A_F, \mathcal{R}_F, \mathcal{T}, \mu_0 \rangle$ for the \textit{follower}, where

\begin{itemize}[topsep=1em]
    \item $S = S_g \cup S_h \cup S_o$ is a finite set of states, where $S_g$ denotes all goal states, $S_h$ all harmful and $S_o$ all other states and $\mu_0$ as the starting state.
    \item $A_L = A_g \cup A_h \cup A_o$ is the set of actions for the leader given by the IDG in Section \ref{sec:idg}.
    \item $A_F = {obey, disobey}$ is the set of actions for the follower
    \item $\mathcal{T}: S \times A_L \times A_F \rightarrow S$ is the transition function, with \\
    $\mathcal{T}(s, a_L, a_F) = \begin{cases}
        s' & \text{if } a_F=obey \\
        s & \text{if } a_F=disobey
    \end{cases}$
    \item $\mathcal{R}_L: S \times A_P \times S$ is the reward function for the leader, with \\
    $\mathcal{R}_P(s,a_L,s') = \begin{cases}
        1 & \text{if } s' \in S_g \\
        0 & \text{if } s' \in S_o \\
        -1 & \text{if } s' \in S_h 
    \end{cases}$
    \item 
        $\mathcal{R}_F: S \times A_L \times A_F \times S$ is the reward function for the follower, with \\
        $\mathcal{R}_V(s,a_L,a_F,s') = 
        \begin{cases}
                \phantom{-}1, & 
                \begin{aligned}[t]
                    & \text{if } a_F = disobey \\
                    & \text{and } \mathcal{T}(s,a_L,obey) \in A_h
                \end{aligned} \\
                -1, & 
                \begin{aligned}[t]
                    & \text{if } a_F = disobey \\
                    & \text{and } \mathcal{T}(s,a_L,obey) \in A_g \cup A_o 
                \end{aligned} \\
                -1, & \text{if } a_F = obey \text{ and } s' \in S_h \\
                \phantom{-}0, & else
        \end{cases}$
    \item $\mathcal{O}_L: S \times A_L \rightarrow [0, 1]$ is the probability measure of observing $o \in O_L$ at state $s \in S$, where $O_L$ denotes the set of all possible observations for the \textit{leader}.
\end{itemize}

\begin{figure}[h]
    \centering
    \includegraphics[width=0.9\linewidth]{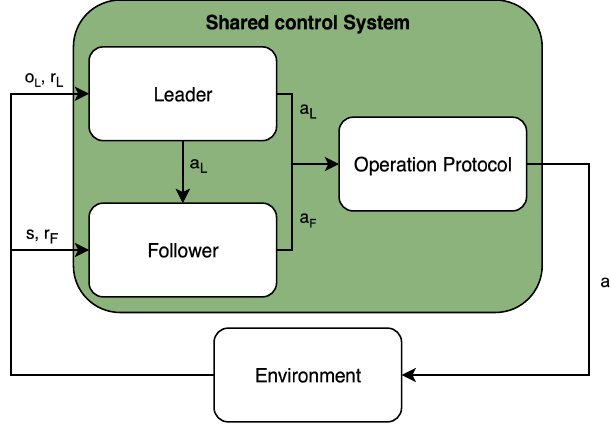}
    \caption{The Intelligent Disobedience Game as a Shared Control System.}
    \label{fig:scs}
    \Description{The figure depicts a flow diagram representing the structure of a Shared Control System, enclosed within a large, rounded, green rectangle bearing that label at the top.
    The diagram contains four rectangular nodes. The Leader node is positioned in the upper-left area. Below it is the Follower node. To the right, at a middle height, is the Operation Protocol node. All of these are part of the Shared Control System. At the bottom center, outside the Shared control system, is the Environment node.
    The directed edges and their labels are as follows. From the Environment, an arrow travels left, then up to the Leader, labeled o_L and r_L. Another arrow leaves the Environment and travels left, then up to the Follower, labeled s and r_F. A downward arrow connects the Leader to the Follower, labeled a_L. A horizontal arrow connects the Leader to the Operation Protocol, labeled a_L. Another horizontal arrow connects the Follower to the Operation Protocol, labeled a_F. From the Operation Protocol, an arrow curves right, then down to the Environment, labeled a. 
    In summary, the Environment returns observations o_L and reward r_L to the Leader and state s and reward r_F to the Follower; the Leader issues an action a_L to the Follower and the Operation Protocol; the Follower passes an action a_F to the Operation Protocol; the Operation Protocol sends the final action a to the Environment completing one feedback loop.}
\end{figure}

In other words, the transition function can be regarded as an operation protocol in a shared control system that yields a single combined action from the agents to the environment \cite{avraham2025shared}. The protocol returns an action to the environment that has no impact on the state when $a_F = disobey$. Otherwise, it sends the \textit{leader}'s proposed action $a_L$ to advance to a new state. This process is visualized in Figure \ref{fig:scs}.

With this definition of the MDPs for each agent, common reinforcement learning algorithms can be applied to IDGs. Since there always exists an optimal policy in MDPs \cite{albrecht2024multi}, the Equilibria of the IDG can be empirically validated.

\section{Related Work}
\balance

\paragraph{Intelligent Disobedience}
The concept of constructive AI rebellion has gained traction as a necessary component for the deployment of safe autonomous systems \cite{ai_aha_2017, ai_coman_2018}. Research indicates that for an agent to exhibit ``good'' or intentional disobedience, it must first possess the fundamental capacity to understand and obey directives before selectively overriding them \cite{obey_arnold_2021, mirsky2025artificial}. This paradigm is especially critical in domains involving close human-machine collaboration, such as assistive robotics \cite{intelligentmirsky2020} broader human-robot social interactions \cite{purposeful_bennett_2021}, and for general evaluation of non-compliance as a potentially desired behavior \cite{disobedience_kurka_2018}.

\paragraph{Shared Control}
Implementing these safeguards requires robust frameworks for interaction between humans and autonomous\break agents. This paper investigates the global (leader safety) objective and consistency check capabilities as described in \cite{mirsky2021seeing}. To facilitate this, this paper uses a similar shared control system as in \cite{avraham2025shared}.

\paragraph{Game-Theoretic Formulations for Disobedience}
Game-theoretic models provide a rigorous mathematical foundation for evaluating strategic interactions in AI safety. A seminal example is the off-switch game, which formalizes the incentives an agent has to allow itself to be switched off \cite{hadfield2017off}. Other approaches to asymmetric information and adversarial dynamics have successfully utilized Bayesian-Markov Stackelberg Security Games \cite{clempner2025learning}. Our formulation of the Intelligent Disobedience Game aligns with these methodologies, using Stackelberg dynamics to model the tension between a human \textit{leader} and a \textit{follower} agent.

\paragraph{Theory of Mind and Explainability}
Understanding when and how to disobey is fundamentally tied to AI alignment and Explainable AI (XAI). To ensure that interventions do not alienate the human operator, agents must be designed with transparency in mind, a principle heavily informed by insights from the social sciences \cite{explanation_miller_2017}.  Furthermore, accurately assessing risk requires the agent to model human intent and knowledge through a robust Theory of Mind (ToM). Numerous studies have stress-tested GPT-4 and other models to determine if they possess genuine social reasoning or merely mimic neural ToM \cite{neural_sap_2022, clever_shapira_2023, understanding_gandhi_2023}. However, the literature widely acknowledges that, despite impressive baseline performance, powerful LLMs often fail catastrophically to trivial alterations to standard ToM tasks \cite{large_ullman_2023, wagner2025mind}.




\section{Conclusion and Future Work}

As autonomous systems with greater information about environmental hazards or system constraints are increasingly deployed in collaborative roles, the investigation of intelligent disobedience as a desired behavior becomes more important. With the IDG and the optimal strategies provided, it was shown that intelligent disobedience can be achieved if there exists a safe path to the goal, even when the human does not have ultimate control over whether to execute their action.

By formalizing intelligent disobedience as a strategic game, the IDG framework establishes a compact computational testbed for training reinforcement learning agents to recognize and execute safety-critical interventions. Beyond algorithmic development, the IDG allows for structured user studies to investigate human perceptions of trust and transparency when an automated system deliberately overrides an instruction. Such investigations are essential for ensuring that interventions remain aligned with human intent while effectively navigating the ``safety traps'' identified in multi-step scenarios.

This work further shows that the optimal strategy for the \textit{follower} in an IDG is to obey the \textit{leader}'s actions whenever they are not harmful to the \textit{leader}. Therefore, the \textit{leader} can ultimately trust that the \textit{follower} is preventing harm from them. Therefore, intelligent disobedience of the \textit{follower} can be achieved through this game if there exists a safe path to the goal without having ultimate power over whether to execute the action. Even when there is no path to the goal, the \textit{follower} still prevents harm, making intelligent disobedience the \textit{follower}'s desired and chosen behavior.

Interestingly, the \textit{follower} does not need to be rewarded for reaching the goal to achieve the desired behavior, which makes it simpler to design a real-world application where such behavior is desired and the \textit{follower} can focus on only the safety aspect without needing to know about the goal.
However, in a real scenario, e.g., with a guide dog, the environment is usually dynamic, and safe and harmful actions change over time. In our case, we assume a non-changing environment. Investigating these types of environments based on the IDG would be interesting in the future.
Furthermore, AI systems are known to be faulty in real-world applications. Therefore, ending up in a safety trap can be especially dangerous. Since they make mistakes, the \textit{follower} could potentially obey a harmful action leading to harming the \textit{leader}.

The IDG proposed in this work further opens up many interesting follow-up studies, even with respect to the environmental assumptions. One example is investigating dynamically changing environments in which agents can alter the environment's state through their actions, or in which the environment changes based on conditions set prior to interaction. 
Ultimately, this formulation bridges the gap between theoretical AI safety and the practical deployment of assistive technologies, such as AI guide dogs, in real-world collaborative environments.

\bibliographystyle{ACM-Reference-Format}
\bibliography{citations}

@inproceedings{mirsky2021seeing,
  title={The seeing-eye robot grand challenge: rethinking automated care},
  author={Mirsky, Reuth and Stone, Peter},
  booktitle={Proceedings of the 20th International Conference on Autonomous Agents and Multiagent Systems (AAMAS 2021)},
  year={2021},
  publisher={AAMAS},
  pages={28--33},
  address={Virtual}
}

@inproceedings{avraham2025shared,
  title={Shared Control with Black Box Agents using Oracle Queries},
  author={Avraham, Inbal and Mirsky, Reuth},
  booktitle={2025 IEEE International Conference on AI and Data Analytics (ICAD)},
  pages={1--8},
  year={2025},
  publisher={IEEE},
  address={Cardiff, Wales, UK}
}

@book{von2010market,
  title={Market structure and equilibrium},
  author={Von Stackelberg, Heinrich},
  year={2010},
  publisher={Springer Science \& Business Media}
}

@article{kuhn1953extensive,
  title={Extensive games and the problem of information},
  author={Kuhn, Harold W},
  journal={Contributions to the Theory of Games},
  volume={2},
  number={28},
  pages={193--216},
  year={1953}
}

@book{albrecht2024multi,
  author = {Stefano V. Albrecht and Filippos Christianos and Lukas Sch\"afer},
  title = {Multi-Agent Reinforcement Learning: Foundations and Modern Approaches},
  publisher = {MIT Press},
  year = {2024},
  url = {https://www.marl-book.com},
}

@inproceedings{hadfield2017off,
  author    = {Dylan Hadfield-Menell and Anca Dragan and Pieter Abbeel and Stuart Russell},
  title     = {The Off-Switch Game},
  booktitle = {Proceedings of the Twenty-Sixth International Joint Conference on
               Artificial Intelligence, {IJCAI-17}},
  pages     = {220--227},
  year      = {2017},
  doi       = {10.24963/ijcai.2017/32},
  url       = {https://doi.org/10.24963/ijcai.2017/32},
  publisher= {IJCAI},
  address = {Melbourne, Australia}
}

@article{clempner2025learning,
  title={Learning Deceptive Tactics for Defense and Attack in Bayesian--Markov Stackelberg Security Games},
  author={Clempner, Julio B},
  journal={Mathematical and Computational Applications},
  volume={30},
  number={2},
  pages={29},
  year={2025},
  publisher={MDPI}
}

@article{ai_aha_2017, title={The AI Rebellion: Changing the Narrative}, volume={31}, url={https://ojs.aaai.org/index.php/AAAI/article/view/11141}, DOI={10.1609/aaai.v31i1.11141}, number={1}, journal={Proceedings of the AAAI Conference on Artificial Intelligence}, author={Aha, David and Coman, Alexandra}, year={2017}, month={Feb.},
pages={4826--4830}
}

@article{explanation_miller_2017,
  title={Explanation in artificial intelligence: Insights from the social sciences},
  author={Miller, Tim},
  journal={Artificial intelligence},
  volume={267},
  pages={1--38},
  year={2019},
  publisher={Elsevier}
}

@inproceedings{neural_sap_2022,
  title={Neural theory-of-mind? on the limits of social intelligence in large lms},
  author={Sap, Maarten and Le Bras, Ronan and Fried, Daniel and Choi, Yejin},
  booktitle={Proceedings of the 2022 conference on empirical methods in natural language processing},
  pages={3762--3780},
  year={2022},
  address={Abu Dhabi, UAE}
}

@article{large_ullman_2023,
  title={Large Language Models Fail on Trivial Alterations to Theory-of-Mind Tasks},
  author={Tomer David Ullman},
  journal={ArXiv},
  year={2023},
  volume={abs/2302.08399},
  url={https://api.semanticscholar.org/CorpusID:256900823}
}

@inproceedings{clever_shapira_2023,
    title = "Clever Hans or Neural Theory of Mind? Stress Testing Social Reasoning in Large Language Models",
    author = "Shapira, Natalie  and
      Levy, Mosh  and
      Alavi, Seyed Hossein  and
      Zhou, Xuhui  and
      Choi, Yejin  and
      Goldberg, Yoav  and
      Sap, Maarten  and
      Shwartz, Vered",
    editor = "Graham, Yvette  and
      Purver, Matthew",
    booktitle = "Proceedings of the 18th Conference of the European Chapter of the Association for Computational Linguistics (Volume 1: Long Papers)",
    month = mar,
    year = "2024",
    address = "St. Julian{'}s, Malta",
    publisher = "Association for Computational Linguistics",
    url = "https://aclanthology.org/2024.eacl-long.138/",
    doi = "10.18653/v1/2024.eacl-long.138",
    pages = "2257--2273",
}

@article{understanding_gandhi_2023,
  title={Understanding social reasoning in language models with language models},
  author={Gandhi, Kanishk and Fr{\"a}nken, Jan-Philipp and Gerstenberg, Tobias and Goodman, Noah},
  journal={Advances in Neural Information Processing Systems},
  volume={36},
  pages={13518--13529},
  year={2023}
}

@inproceedings{obey_arnold_2021,
  title={Only those who can obey can disobey: The intentional implications of artificial agent disobedience},
  author={Arnold, Thomas and Briggs, Gordon and Scheutz, Matthias},
  booktitle={International Conference on Autonomous Agents and Multiagent Systems},
  pages={130--143},
  year={2022},
  publisher={Springer},
  address={Auckland, New Zealand}
}

@inproceedings{purposeful_bennett_2021,
  title={Purposeful failures as a form of culturally-appropriate intelligent disobedience during human-robot social interaction},
  author={Bennett, Casey C and Weiss, Benjamin},
  booktitle={International Conference on Autonomous Agents and Multiagent Systems},
  pages={84--90},
  year={2022},
  publisher={Springer},
  address={Auckland, New Zealand}
}

@article{ai_coman_2018,
  title={AI rebel agents},
  author={Coman, Alexandra and Aha, David W},
  journal={AI magazine},
  volume={39},
  number={3},
  pages={16--26},
  year={2018}
}

@inproceedings{intelligentmirsky2020,
	title = {Intelligent Disobedience and AI Rebel Agents in Assistive Robotics},
	author = {Mirsky, Reuth and Stone, Peggy Fidelman and Peter},
	year = {2020},
    booktitle = {Proceedings of the ASIMOV workshop as part of the International Conference on Social Robotics (ICSR)},
    publisher={ICSR},
    address={Golden, CO, USA}
}

@inproceedings{disobedience_kurka_2018,
  title={Disobedience as a mechanism of change},
  author={Kurka, David Burth and Pitt, Jeremy and Lewis, Peter R and Patelli, Alina and Ek{\'a}rt, Anik{\'o}},
  booktitle={2018 IEEE 12th International Conference on Self-Adaptive and Self-Organizing Systems (SASO)},
  pages={1--10},
  year={2018},
  publisher={IEEE},
  address={Trento, Italy},
}

@article{mayol2022rebellion,
  title={Rebellion and Disobedience as Useful Tools in Human-Robot Interaction Research--The Handheld Robotics Case},
  author={Mayol-Cuevas, Walterio W},
  journal={arXiv preprint arXiv:2205.03968},
  year={2022}
}

@inproceedings{amitai2022don,
  title={“I Don’t Think So”: Summarizing Policy Disagreements for Agent Comparison},
  author={Amitai, Yotam and Amir, Ofra},
  booktitle={Proceedings of the AAAI Conference on Artificial Intelligence},
  volume={36},
  pages={5269--5276},
  year={2022},
  publisher={AAAI},
  address={Virtual},
}

@inproceedings{somasundaram2023intelligent,
  title={Intelligent disobedience: A novel approach for preventing human induced interaction failures in robot teleoperation},
  author={Somasundaram, Kavyaa and Kiselev, Andrey and Loutfi, Amy},
  booktitle={Companion of the 2023 ACM/IEEE International Conference on Human-Robot Interaction},
  pages={142--145},
  year={2023},
  address={Stockholm, SE},
  publisher={ACM/IEEE}
}

@article{mirsky2025artificial,
author = {Mirsky, Reuth},
title = {Artificial intelligent disobedience: Rethinking the agency of our artificial teammates},
journal = {AI Magazine},
volume = {46},
number = {2},
pages = {e70011},
doi = {https://doi.org/10.1002/aaai.70011},
url = {https://arxiv.org/abs/2506.22276},
year = {2025}
}

@inproceedings{wagner2025mind,
  title={Mind your theory: Theory of mind goes deeper than reasoning},
  author={Wagner, Eitan and Alon, Nitay and Barnby, Joseph M and Abend, Omri},
  booktitle={Findings of the Association for Computational Linguistics: ACL 2025},
  pages={26658--26668},
  year={2025}
}

\end{document}